# AN EXTENDED THEORY OF HEAD-DRIVEN PARSING

**Mark-Jan Nederhof** *
University of Nijmegen
Department of Computer Science
Toernooiveld, 6525 ED Nijmegen
The Netherlands
markjan@cs.kun.nl

**Giorgio Satta**
Università di Padova
Dipartimento di Elettronica e Informatica
via Gradenigo 6/A, 35131 Padova
Italy
satta@dei.unipd.it

## Abstract

We show that more head-driven parsing algorithms can be formulated than those occurring in the existing literature. These algorithms are inspired by a family of left-to-right parsing algorithms from a recent publication. We further introduce a more advanced notion of "head-driven parsing" which allows more detailed specification of the processing order of non-head elements in the right-hand side. We develop a parsing algorithm for this strategy, based on LR parsing techniques.

## Introduction

According to the head-driven paradigm, parsing of a formal language is started from the elements within the input string that are most contentful either from a syntactic or, more generally, from an information theoretic point of view. This results in the weakening of the left-to-right feature of most traditional parsing methods. Following a pervasive trend in modern theories of Grammar (consider for instance [5, 3, 11]) the computational linguistics community has paid large attention to the head-driven paradigm by investigating its applications to context-free language parsing.

Several methods have been proposed so far exploiting some nondeterministic head-driven strategy for context-free language parsing (see among others [6, 13, 2, 14]). All these proposals can be seen as generalizations to the head-driven case of parsing prescriptions originally conceived for the left-to-right case. The methods above suffer from deficiencies that are also noticeable in the left-to-right case. In fact, when more rules in the grammar share the same head element, or share some infix of their right-hand side including the head, the recognizer nondeterministically guesses a rule just after having seen the head. In this way analyses that could have been shared are duplicated in the parsing process.

Interesting techniques have been proposed in the left-to-right deterministic parsing literature to overcome redundancy problems of the above kind, thus reducing the degree of nondeterminism of the resulting methods. These solutions range from predictive LR parsing to LR parsing [15, 1]. On the basis of work in [8] for nondeterministic left-to-right parsing, we trace here a theory of head-driven parsing going from crude top-down and head-corner to more sophisticated solutions, in the attempt to successively make more deterministic the behaviour of head-driven methods.

Finally, we propose an original generalization of head-driven parsing, allowing a more detailed specification of the order in which elements of a right-hand side are to be processed. We study in detail a solution to such a head-driven strategy based on LR parsing. Other methods presented in this paper could be extended as well.

## Preliminaries

The notation used in the sequel is for the most part standard and is summarised below.

Let $D$ be an alphabet (a finite set of symbols); $D^+$ denotes the set of all (finite) non-empty strings over $D$ and $D^*$ denotes $D^+ \cup \{\varepsilon\}$, where $\varepsilon$ denotes the empty string. Let $R$ be a binary relation; $R^+$ denotes the transitive closure of $R$ and $R^*$ denotes the reflexive and transitive closure of $R$.

A context-free grammar $G = (N, T, P, S)$ consists of two finite disjoint sets $N$ and $T$ of nonterminal and terminal symbols, respectively, a start symbol $S \in N$, and a finite set of rules $P$. Every rule has the form $A \to \alpha$, where the left-hand side (lhs) $A$ is an element from $N$ and the right-hand side (rhs) $\alpha$ is an element from $V^+$, where $V$ denotes $(N \cup T)$. (Note that we do not allow rules with empty right-hand sides. This is for the sake of presentational simplicity.) We use symbols $A, B, C, \ldots$ to range over $N$, symbols $X, Y, Z$ to range over $V$, symbols $\alpha, \beta, \gamma, \ldots$ to range over $V^*$, and $v, w, x, \ldots$ to range over $T^*$.

In the context-free grammars that we will consider, called *head grammars*, exactly one member from each rhs is distinguished as the *head*. We indicate the head by underlining it, e.g., we write $A \to \alpha \underline{X} \beta$. An expression $A \to \alpha \overline{\gamma} \beta$ denotes a rule in which the head is some member within $\gamma$. We define a binary relation $\diamond$ such that $B \diamond A$ if and only if $A \to \alpha \underline{B} \beta$ for some $\alpha$ and $\beta$. Relation $\diamond^*$ is called the *head-corner relation*.

---

*Supported by the Dutch Organisation for Scientific Research (NWO), under grant 00-62-518

For technical reasons we sometimes need the *augmented* set of rules $P^\dagger$, consisting of all rules in $P$ plus the extra rule $S' \to \bot S$, where $S'$ is a fresh nonterminal, and $\bot$ is a fresh terminal acting as an imaginary zeroth input symbol. The relation $P^\dagger$ is extended to a relation $\to$ on $V^* \times V^*$ as usual. We write $\gamma \xrightarrow{p} \delta$ whenever $\gamma \to \delta$ holds as an extension of $p \in P^\dagger$. We write $\gamma \xrightarrow{p_1 p_2 \cdots p_s} \delta$ if $\gamma \xrightarrow{p_1} \delta_1 \xrightarrow{p_2} \delta_2 \cdots \delta_{s-1} \xrightarrow{p_s} \delta$

For a fixed grammar, a head-driven recognition algorithm can be specified by means of a stack automaton $A = (T, Alph, Init(n), \mapsto, Fin(n))$, parameterised with the length $n$ of the input. In $A$, symbols $T$ and $Alph$ are the input and stack alphabets respectively, $Init(n), Fin(n) \in Alph$ are two distinguished stack symbols and $\mapsto$ is the transition relation, defined on $Alph^+ \times Alph^+$ and implicitly parameterised with the input.

Such an automaton manipulates stacks $\Gamma \in Alph^+$, (constructed from left to right) while consulting the symbols in the given input string. The initial stack is $Init(n)$. Whenever $\Gamma \mapsto \Gamma'$ holds, one step of the automaton may, under some conditions on the input, transform a stack of the form $\Gamma'' \Gamma$ into the stack $\Gamma'' \Gamma'$. In words, $\Gamma \mapsto \Gamma'$ denotes that if the top-most few symbols on the stack are $\Gamma$ then these may be replaced by the symbols $\Gamma'$. Finally, the input is accepted whenever the automaton reaches stack $Fin(n)$. Stack automata presented in what follows act as recognizers. Parsing algorithms can directly be obtained by pairing these automata with an output effect.

## A family of head-driven algorithms

This section investigates the adaptation of a family of left-to-right parsing algorithms from [8], viz. top-down, left-corner, PLR, ELR, and LR parsing, to head grammars.

### Top-down parsing

The following is a straightforward adaptation of top-down (TD) parsing [1] to head grammars.

There are two kinds of stack symbol (*items*), one of the form $[i, A, j]$, which indicates that some subderivation from $A$ is needed deriving a substring of $a_{i+1} \ldots a_j$, the other of the form $[i, k, A \to \alpha \bullet \gamma \bullet \beta, m, j]$, which also indicates that some subderivation from $A$ is needed deriving a substring of $a_{i+1} \ldots a_j$, but specifically using the rule $A \to \alpha \gamma \beta$, where $\gamma \to^* a_{k+1} \ldots a_m$ has already been established. Formally, we have

$$I_1^{TD} = \{[i, A, j] \mid i < j\}$$
$$I_2^{TD} = \{[i, k, A \to \alpha \bullet \gamma \bullet \beta, m, j] \mid A \to \alpha \underline{\gamma} \beta \in P^\dagger \wedge i \leq k \leq m \leq j\}$$

**Algorithm 1 (Head-driven top-down)**
$A^{TD} = (T, I_1^{TD} \cup I_2^{TD}, Init(n), \mapsto, Fin(n))$, where
$Init(n) = [-1, -1, S' \to \bullet \bot \bullet S, 0, n]$,
$Fin(n) = [-1, -1, S' \to \bullet \bot S \bullet, n, n]$, and the transition relation $\mapsto$ is given by the following clauses.

0  $[i, A, j] \mapsto [i, A, j][i, B, j]$
   where there is $A \to \alpha \underline{B} \beta \in P^\dagger$
0a $[i, k, A \to \alpha \bullet \gamma \bullet B\beta, m, j] \mapsto$
   $[i, k, A \to \alpha \bullet \gamma \bullet B\beta, m, j][m, B, j]$
0b $[i, k, A \to \alpha B \bullet \gamma \bullet \beta, m, j] \mapsto$
   $[i, k, A \to \alpha B \bullet \gamma \bullet \beta, m, j][i, B, k]$
1  $[i, A, j] \mapsto [i, k-1, A \to \alpha \bullet a \bullet \beta, k, j]$
   where there are $A \to \alpha \underline{a} \beta \in P^\dagger$ and $k$ such that $i < k \leq j$ and $a_k = a$
2a $[i, k, A \to \alpha \bullet \gamma \bullet a\beta, m, j] \mapsto$
   $[i, k, A \to \alpha \bullet \gamma a \bullet \beta, m+1, j]$
   provided $m < j$ and $a_{m+1} = a$
2b Symmetric to 2a (cf. 0a and 0b)
3  $[i, A, j][i', k, B \to \bullet \delta \bullet, m, j'] \mapsto$
   $[i, k, A \to \alpha \bullet B \bullet \beta, m, j]$
   where there is $A \to \alpha \underline{B} \beta \in P^\dagger$
   ($i = i'$ and $j = j'$ are automatically satisfied)
4a $[i, k, A \to \alpha \bullet \gamma \bullet B\beta, m, j][i', k', B \to \bullet \delta \bullet, m', j'] \mapsto$
   $[i, k, A \to \alpha \bullet \gamma B \bullet \beta, m', j]$
   provided $m = k'$
   ($m = i'$ and $j = j'$ are automatically satisfied)
4b Symmetric to 4a

We call a grammar *head-recursive* if $A \diamond^+ A$ for some $A$. Head-driven TD parsing may loop exactly for the grammars which are head-recursive. Head recursion is a generalization of left recursion for traditional TD parsing.

In the case of grammars with some parameter mechanism, top-down parsing has the advantage over other kinds of parsing that top-down propagation of parameter values is possible in collaboration with context-free parsing (cf. the standard evaluation of definite clause grammars), which may lead to more efficient processing. This holds for left-to-right parsing as well as for head-driven parsing [10].

### Head-corner parsing

The predictive steps from Algorithm 1, represented by Clause 0 and supported by Clauses 0a and 0b, can be compiled into the head-corner relation $\diamond^*$. This gives the head-corner (HC) algorithm below. The items from $I_1^{TD}$ are no longer needed now. We define $I^{HC} = I_2^{TD}$.

**Algorithm 2 (head-corner)**
$A^{HC} = (T, I^{HC}, Init(n), \mapsto, Fin(n))$, where
$Init(n) = [-1, -1, S' \to \bullet \bot \bullet S, 0, n]$,
$Fin(n) = [-1, -1, S' \to \bullet \bot S \bullet, n, n]$, and $\mapsto$ is given by the following clauses. (Clauses 1b, 2b, 3b, 4b are omitted, since these are symmetric to 1a, 2a, 3a, 4a, respectively.)

1a $[i, k, A \to \alpha \bullet \gamma \bullet B\beta, m, j] \mapsto$
   $[i, k, A \to \alpha \bullet \gamma \bullet B\beta, m, j][m, p-1, C \to \eta \bullet a \bullet \theta, p, j]$
   where there are $C \to \eta \underline{a} \theta \in P^\dagger$ and $p$ such that $m < p \leq j$ and $a_p = a$ and $C \diamond^* B$

2a $[i, k, A \to \alpha \bullet \gamma \bullet a\beta, m, j] \mapsto$
   $[i, k, A \to \alpha \bullet \gamma a \bullet \beta, m+1, j]$
   provided $m < j$ and $a_{m+1} = a$

3a $[i, k, D \to \alpha \bullet \gamma \bullet A\beta, m, j][i', k', B \to \bullet \delta \bullet, m', j'] \mapsto$
   $[i, k, D \to \alpha \bullet \gamma \bullet A\beta, m, j][i', k', C \to \eta \bullet B \bullet \theta, m', j']$
   provided $m = i'$, where there is $C \to \eta \underline{B} \theta \in P^\dagger$ such that $C \diamond^* A$
   ($j = j'$ is automatically satisfied)

4a $[i, k, A \to \alpha \bullet \gamma \bullet B\beta, m, j][i', k', B \to \bullet \delta \bullet, m', j'] \mapsto$
   $[i, k, A \to \alpha \bullet \gamma B \bullet \beta, m', j]$
   provided $m = k'$
   ($m = i'$ and $j = j'$ are automatically satisfied)

Head-corner parsing as well as all algorithms in the remainder of this paper may loop exactly for the grammars which are cyclic (where $A \to^+ A$ for some $A$).

The head-corner algorithm above is the only one in this paper which has already appeared in the literature, in different guises [6, 13, 2, 14].

## Predictive HI parsing

We say two rules $A \to \alpha_1$ and $B \to \alpha_2$ have a *common infix* $\alpha$ if $\alpha_1 = \beta_1 \underline{\alpha} \gamma_1$ and $\alpha_2 = \beta_2 \underline{\alpha} \gamma_2$, for some $\beta_1$, $\beta_2$, $\gamma_1$ and $\gamma_2$. The notion of common infix is an adaptation of the notion of common *prefix* [8] to head grammars.

If a grammar contains many common infixes, then HC parsing may be very nondeterministic; in particular, Clauses 1 or 3 may be applied with different rules $C \to \eta \underline{a} \theta \in P^\dagger$ or $C \to \eta \underline{B} \theta \in P^\dagger$ for fixed $a$ or $B$.

In [15] an idea is described that allows reduction of nondeterminism in case of common prefixes and left-corner parsing. The resulting algorithm is called *predictive LR (PLR) parsing*. The following is an adaptation of this idea to HC parsing. The resulting algorithm is called *predictive HI (PHI) parsing*. (HI parsing, to be discussed later, is a generalization of LR parsing to head grammars.)

First, we need a different kind of item, viz. of the form $[i, k, A \to \gamma, m, j]$, where there is some rule $A \to \alpha \underline{\gamma} \beta$. With such an item, we simulate computation of different items $[i, k, A \to \alpha \bullet \gamma \bullet \beta, m, j] \in I^{HC}$, for different $\alpha$ and $\beta$, which would be treated individually by an HC parser. Formally, we have

$$I^{PHI} = \{[i, k, A \to \gamma, m, j] \mid A \to \alpha \underline{\gamma} \beta \in P^\dagger \land i \leq k \leq m \leq j\}$$

### Algorithm 3 (Predictive HI)

$A^{PHI} = (T, I^{PHI}, Init(n), \mapsto, Fin(n))$, where
$Init(n) = [-1, -1, S' \to \bot, 0, n]$,
$Fin(n) = [-1, -1, S' \to \bot S, n, n]$, and $\mapsto$ is given by the following (symmetric "b-clauses" omitted).

1a $[i, k, A \to \gamma, m, j] \mapsto$
   $[i, k, A \to \gamma, m, j][m, p-1, C \to a, p, j]$
   where there are $C \to \eta \underline{a} \theta, A \to \alpha \underline{\gamma} B\beta \in P^\dagger$ and $p$ such that $m < p \leq j$ and $a_p = a$ and $C \diamond^* B$

2a $[i, k, A \to \gamma, m, j] \mapsto [i, k, A \to \gamma a, m+1, j]$
   provided $m < j$ and $a_{m+1} = a$, where there is $A \to \alpha \underline{\gamma} a \beta \in P^\dagger$

3a $[i, k, D \to \gamma, m, j][i', k', B \to \delta, m', j'] \mapsto$
   $[i, k, D \to \gamma, m, j][i', k', C \to B, m', j']$
   provided $m = i'$ and $B \to \underline{\delta} \in P^\dagger$, where there are $D \to \alpha \underline{\gamma} A\beta, C \to \eta \underline{B} \theta \in P^\dagger$ such that $C \diamond^* A$

4a $[i, k, A \to \gamma, m, j][i', k', B \to \delta, m', j'] \mapsto$
   $[i, k, A \to \gamma B, m', j]$
   provided $m = k'$ and $B \to \underline{\delta} \in P^\dagger$, where there is $A \to \alpha \underline{\gamma} B \beta \in P^\dagger$

## Extended HI parsing

The PHI algorithm can process simultaneously a common infix $\alpha$ in two different rules $A \to \beta_1 \underline{\alpha} \gamma_1$ and $A \to \beta_2 \underline{\alpha} \gamma_2$, which reduces nondeterminism.

We may however also specify an algorithm which succeeds in simultaneously processing *all* common infixes, irrespective of whether the left-hand sides of the corresponding rules are the same. This algorithm is inspired by *extended LR (ELR) parsing* [12, 7] for extended context-free grammars (where right-hand sides consist of regular expressions over $V$). By analogy, it will be called *extended HI (EHI) parsing*.

This algorithm uses yet another kind of item, viz. of the form $[i, k, \{A_1, A_2, \ldots, A_p\} \to \gamma, m, j]$, where there exists at least one rule $A \to \alpha \underline{\gamma} \beta$ for each $A \in \{A_1, A_2, \ldots, A_p\}$. With such an item, we simulate computation of different items $[i, k, A \to \alpha \bullet \gamma \bullet \beta, m, j] \in I^{HC}$ which would be treated individually by an HC parser. Formally, we have

$$\begin{aligned} I^{EHI} &= \{[i, k, \Delta \to \gamma, m, j] \mid \\ &\quad \emptyset \subset \Delta \subseteq \{A \mid A \to \alpha \underline{\gamma} \beta \in P^\dagger\} \land \\ &\quad i \leq k \leq m \leq j\} \end{aligned}$$

### Algorithm 4 (Extended HI)

$A^{EHI} = (T, I^{EHI}, Init(n), \mapsto, Fin(n))$, where
$Init(n) = [-1, -1, \{S'\} \to \bot, 0, n]$,
$Fin(n) = [-1, -1, \{S'\} \to \bot S, n, n]$, and $\mapsto$ is given by:

1a $[i, k, \Delta \to \gamma, m, j] \mapsto$
   $[i, k, \Delta \to \gamma, m, j][m, p-1, \Delta' \to a, p, j]$
   where there is $p$ such that $m < p \leq j$ and $a_p = a$ and $\Delta' = \{C \mid \exists C \to \eta \underline{a} \theta, A \to \alpha \underline{\gamma} B\beta \in P^\dagger (A \in \Delta \land C \diamond^* B)\}$ is not empty

2a $[i, k, \Delta \to \gamma, m, j] \mapsto [i, k, \Delta' \to \gamma a, m+1, j]$
   provided $m < j$ and $a_{m+1} = a$ and $\Delta' = \{A \in \Delta \mid A \to \alpha \underline{\gamma} a \beta \in P^\dagger\}$ is not empty

3a $[i, k, \Delta \to \gamma, m, j][i', k', \Delta' \to \delta, m', j'] \mapsto$
   $[i, k, \Delta \to \gamma, m, j][i', k', \Delta'' \to B, m', j']$
   provided $m = i'$ and $B \to \underline{\delta} \in P^\dagger$ for some $B \in \Delta'$ such that $\Delta'' = \{C \mid \exists C \to \eta \underline{B} \theta, D \to \alpha \underline{\gamma} A\beta \in P^\dagger (D \in \Delta \land C \diamond^* A)\}$ is not empty

4a $[i, k, \Delta \to \gamma, m, j][i', k', \Delta' \to \delta, m', j'] \mapsto$
   $[i, k, \Delta'' \to \gamma B, m', j]$
   provided $m = k'$ and $B \to \underline{\delta} \in P^\dagger$ for some $B \in \Delta'$ such that $\Delta'' = \{A \in \Delta \mid A \to \alpha \underline{\gamma} B\beta \in P^\dagger\}$ is not empty

This algorithm can be simplified by omitting the sets $\Delta$ from the items. This results in *common infix (CI) parsing*, which is a generalization of common prefix parsing [8]. CI parsing does not satisfy the correct subsequence property, to be discussed later. For space reasons, we omit further discussion of CI parsing.

## HI parsing

If we translate the difference between ELR and LR parsing [8] to head-driven parsing, we are led to *HI parsing*, starting from EHI parsing, as described below. The algorithm is called HI because it computes head-inward derivations in reverse, in the same way as LR parsing computes rightmost derivations in reverse [1]. Head-inward derivations will be discussed later in this paper.

HI parsing uses items of the form $[i, k, Q, m, j]$, where $Q$ is a non-empty set of "double-dotted" rules $A \to \alpha \bullet \gamma \bullet \beta$. The fundamental difference with the items in $I^{EHI}$ is that the infix $\gamma$ in the right-hand sides does not have to be fixed. Formally, we have

$$I^{HI} = \{[i, k, Q, m, j] \mid$$
$$\emptyset \subset Q \subseteq \{A \to \alpha \bullet \gamma \bullet \beta \mid A \to \alpha\underline{\gamma}\beta \in P^\dagger\} \wedge$$
$$i \leq k < m \leq j\}$$

We explain the difference in behaviour of HI parsing with regard to EHI parsing by investigating Clauses 1a and 2a of Algorithm 4. (Clauses 3a and 4a would give rise to a similar discussion.) Clauses 1a and 2a both address some terminal $a_p$, with $m < p \leq j$. In Clause 1a, the case is treated that $a_p$ is the head (which is not necessarily the leftmost member) of a rhs which the algorithm sets out to recognize; in Clause 2a, the case is treated that $a_p$ is the next member of a rhs of which some members have already been recognized, in which case we must of course have $p = m + 1$.

By using the items from $I^{HI}$ we may do both kinds of action simultaneously, provided $p = m + 1$ and $a_p$ is the leftmost member of some rhs of some rule, where it occurs as head.[1] The lhs of such a rule should satisfy a requirement which is more specific than the usual requirement with regard to the head-corner relation.[2] We define the left head-corner relation (and the right head-corner relation, by symmetry) as a subrelation of the head-corner relation as follows.

We define: $B \angle A$ if and only if $A \to \underline{B}\alpha$ for some $\alpha$. The relation $\angle^*$ now is called the *left head-corner relation*.

We define

$gotoright_1(Q, X) =$
$\quad \{C \to \eta \bullet X \bullet \theta \mid C \to \eta\underline{X}\theta \in P^\dagger \wedge$
$\quad\quad \exists A \to \alpha \bullet \gamma \bullet B\beta \in Q(C \diamondsuit^* B)\}$

$gotoright_2(Q, X) =$
$\quad \{C \to \bullet X \bullet \theta \mid C \to \underline{X}\theta \in P^\dagger \wedge$
$\quad\quad \exists A \to \alpha \bullet \gamma \bullet B\beta \in Q(C \angle^* B)\} \cup$
$\quad \{A \to \alpha \bullet \gamma X \bullet \beta \mid A \to \alpha \bullet \gamma \bullet X\beta \in Q\}$

and assume symmetric definitions for $gotoleft_1$ and $gotoleft_2$.

The above discussion gives rise to the new Clauses 1a and 2a of the algorithm below. The other clauses are derived analogously from the corresponding clauses of Algorithm 4. Note that in Clauses 2a and 4a the new item does not replace the existing item, but is pushed on top of it; this requires extra items to be popped off the stack in Clauses 3a and 4a.[3]

**Algorithm 5 (HI)**
$A^{HI} = (T, I^{HI}, Init(n), \mapsto, Fin(n))$, where
$Init(n) = [-1, -1, \{S' \to \bullet \bot \bullet S\}, 0, n]$,
$Fin(n) = [-1, -1, \{S' \to \bullet \bot S \bullet\}, n, n]$, and $\mapsto$ defined:

1a $[i, k, Q, m, j] \mapsto [i, k, Q, m, j][m, p-1, Q', p, j]$
where there is $p$ such that $m + 1 < p \leq j$ and $a_p = a$ and $Q' = gotoright_1(Q, a)$ is not empty

2a $[i, k, Q, m, j] \mapsto [i, k, Q, m, j][i, k, Q', m+1, j]$
provided $m < j$ and $a_{m+1} = a$ and $Q' = gotoright_2(Q, a)$ is not empty

3a $[i, k, Q, m, j]I_1 \ldots I_{r-1}[i', k', Q', m', j'] \mapsto$
$[i, k, Q, m, j][i', k', Q'', m', j']$
provided $m < k'$, where there is $B \to \bullet X_1 \ldots X_r \bullet$
$\in Q'$ such that $Q'' = gotoright_1(Q, B)$ is not empty

4a $[i, k, Q, m, j]I_1 \ldots I_{r-1}[i', k', Q', m', j'] \mapsto$
$[i, k, Q, m, j][i, k, Q'', m', j]$
provided $m = k'$ or $k = k'$, where there is $B \to \bullet X_1 \ldots X_r \bullet \in Q'$ such that $Q'' = gotoright_2(Q, B)$ is not empty

We feel that this algorithm has only limited advantages over the EHI algorithm for other than degenerate head grammars, in which the heads occur either mostly leftmost or mostly rightmost in right-hand sides. In particular, if there are few sequences of rules of the form $A \to \underline{A_1}\alpha_1, A_1 \to \underline{A_2}\alpha_2, \ldots, A_{m-1} \to \underline{A_m}\alpha_m$, or of the form $A \to \alpha_1\underline{A_1}, A_1 \to \alpha_2\underline{A_2}, \ldots, A_{m-1} \to \alpha_m\underline{A_m}$, then the left and right head-corner relations are very sparse and HI parsing virtually simplifies to EHI parsing.

In the following we discuss a variant of head grammars which may provide more opportunities to use the advantages of the LR technique.

## A generalization of head grammars

The essence of head-driven parsing is that there is a distinguished member in each rhs which is recognized first. Subsequently, the other members to the right and to the left of the head may be recognized.

An artifact of most head-driven parsing algorithms is that the members to the left of the head are recognized strictly from right to left, and vice versa for the members to the right of the head (although recognition of the members in the left part and in the right part may be interleaved). This restriction does not seem to be justified, except by some practical considerations, and it prevents truly non-directional parsing.

---

[1] If $a_p$ is not the leftmost member, then no successful parse will be found, due to the absence of rules with empty right-hand sides (*epsilon rules*).

[2] Again, the absence of epsilon rules is of importance here.

[3] $I_1 \ldots I_{r-1}$ represent a number of items, as many as there are members in the rule recognized, minus one.

We propose a generalization of head grammars in such a way that each of the two parts of a rhs on both sides of the head again have a head. The same holds recursively for the smaller parts of the rhs. The consequence is that a rhs can be seen as a binary tree, in which each node is labelled by a grammar symbol. The root of the tree represents the main head. The left son of the root represents the head of the part of the rhs to the left of the main head, etc.

We denote binary trees using a linear notation. For example, if $\alpha$ and $\beta$ are binary trees, then $(\alpha)X(\beta)$ denotes the binary tree consisting of a root labelled $X$, a left subtree $\alpha$ and a right subtree $\beta$. The notation of empty (sub)trees ($\epsilon$) may be omitted. The relation $\rightarrow^*$ ignores the head information as usual.

Regarding the procedural aspects of grammars, generalized head grammars have no more power than traditional head grammars. This fact is demonstrated by a transformation $\tau_{head}$ from the former to the latter class of grammars. A transformed grammar $\tau_{head}(G)$ contains special nonterminals of the form $[\alpha]$, where $\alpha$ is a proper subtree of some rhs in the original grammar $G = (T, N, P, S)$. The rules of the transformed grammar are given by:

$$A \rightarrow [\alpha]\ \underline{X}\ [\beta] \quad \text{for each } A \rightarrow (\alpha)X(\beta) \in P$$
$$[(\alpha)X(\beta)] \rightarrow [\alpha]\ \underline{X}\ [\beta] \quad \text{for each proper subtree } (\alpha)X(\beta) \text{ of a rhs in } G$$

where we assume that each member of the form $[\epsilon]$ in the transformed grammar is omitted.

It is interesting to note that $\tau_{head}$ is a generalization of a transformation $\tau_{two}$ which can be used to transform a context-free grammar into *two normal form* (each rhs contains one or two symbols). A transformed grammar $\tau_{two}(G)$ contains special nonterminals of the form $[\alpha]$, where $\alpha$ is a proper suffix of a rhs in $G$. The rules of $\tau_{two}(G)$ are given by

$$A \rightarrow X\ [\alpha] \quad \text{for each } A \rightarrow X\alpha \in P$$
$$[X\alpha] \rightarrow X\ [\alpha] \quad \text{for each proper suffix } X\alpha \text{ of a rhs in } G$$

where we assume that each member of the form $[\epsilon]$ in the transformed grammar is omitted.

## HI parsing revisited

Our next step is to show that generalized head grammars can be effectively handled with a generalization of HI parsing (*generalized HI (GHI) parsing*). This new algorithm exhibits a superficial similarity to the 2-dimensional LR parsing algorithm from [16]. For a set Q of trees and rules,[4] $closure(Q)$ is defined to be

---
[4]It is interesting to compare the relation between trees and rules with the one between kernel and nonkernel items of LR parsing [1].

the smallest set which satisfies

$closure(Q) \supseteq Q\ \cup$
$\quad \{A \rightarrow (\alpha)X(\beta) \in P \mid (\gamma)A(\delta) \in closure(Q) \vee$
$\quad\quad\quad\quad\quad\quad\quad\quad\quad\quad B \rightarrow (\gamma)A(\delta) \in closure(Q)\}$

The trees or rules of which the main head is some specified symbol $X$ can be selected from a set $Q$ by

$goto(Q, X) = \{t \in Q \mid t = (\alpha)X(\beta) \vee t = A \rightarrow (\alpha)X(\beta)\}$

In a similar way, we can select trees and rules according to a left or right subtree.

$gotoleft(Q, \alpha) = \{t \in Q \mid t = (\alpha)X(\beta) \vee$
$\quad\quad\quad\quad\quad\quad\quad\quad t = A \rightarrow (\alpha)X(\beta)\}$

We assume a symmetric definition for *gotoright*.

When we set out to recognize the left subtrees from a set of trees and rules, we use the following function.

$left(Q) = closure(\{\alpha \mid (\alpha)X(\beta) \in Q \vee$
$\quad\quad\quad\quad\quad\quad\quad\quad A \rightarrow (\alpha)X(\beta) \in Q\})$

We assume a symmetric definition for *right*.

The set $I^{GHI}$ contains different kinds of item:

- Items of the form $[i, k, Q, m, j]$, with $i \leq k < m \leq j$, indicate that trees $(\alpha)X(\beta)$ and rules $A \rightarrow (\alpha)X(\beta)$ in $Q$ are needed deriving a substring of $a_{i+1} \ldots a_j$, where $X \rightarrow^* a_{k+1} \ldots a_m$ has already been established.

- Items of the form $[k, Q, m, j]$, with $k < m \leq j$, indicate that trees $(\alpha)X(\beta)$ and rules $A \rightarrow (\alpha)X(\beta)$ in $Q$ are needed deriving a substring of $a_{k+1} \ldots a_j$, where $\alpha X \rightarrow^* a_{k+1} \ldots a_m$ has already been established.

  Items of the form $[i, k, Q, m]$ have a symmetric meaning.

- Items of the form $[k, t, m]$, with $k < m$, indicate that $\gamma \rightarrow^* a_{k+1} \ldots a_m$ has been established for tree $t = \gamma$ or rule $t = A \rightarrow \gamma$.

**Algorithm 6 (Generalized HI parsing)**
$A^{GHI} = (T, I^{GHI}, Init(n), \mapsto, Fin(n))$, where
$Init(n) = [-1, \{S' \rightarrow \bot(S)\}, 0, n]$,
$Fin(n) = [-1, S' \rightarrow \bot(S), n]$, and $\mapsto$ defined:

1a $[i, k, Q, m, j] \mapsto [i, k, Q', m]$
  provided $Q' = gotoright(Q, \epsilon)$ is not empty
1b $[i, k, Q, m, j] \mapsto [k, Q', m, j]$
  provided $Q' = gotoleft(Q, \epsilon)$ is not empty
1c $[k, Q, m, j] \mapsto [k, t, m]$
  provided $t \in gotoright(Q, \epsilon)$
1d $[i, k, Q, m] \mapsto [k, t, m]$
  provided $t \in gotoleft(Q, \epsilon)$
2a $[i, k, Q, m, j] \mapsto [i, k, Q, m, j][m, p-1, Q', p, j]$
  where there is $p$ such that $m < p \leq j$ and $Q' = goto(right(Q), a_p)$ is not empty
2b $[i, k, Q, m, j] \mapsto [i, k, Q, m, j][i, p-1, Q', p, k]$
  where there is $p$ such that $i < p \leq k$ and $Q' = goto(left(Q), a_p)$ is not empty

| Stack | Clause |
|---|---|
| $[-1, \{S' \to \bot(S)\}, 0, 4]$ | |
| $[-1, \{S' \to \bot(S)\}, 0, 4]\ [0, 3, \{S \to ((c)A(b))s, S \to (A(d))s, S \to (B)s\}, 4, 4]$ | 3a |
| $[-1, \{S' \to \bot(S)\}, 0, 4]\ [0, 3, \{S \to ((c)A(b))s, S \to (A(d))s, S \to (B)s\}, 4]$ | 1a |
| $[-1, \{S' \to \bot(S)\}, 0, 4]\ [0, 3, \{S \to ((c)A(b))s, S \to (A(d))s, S \to (B)s\}, 4]\ [0, 1, \{A \to a\}, 2, 3]$ | 3b |
| $[-1, \{S' \to \bot(S)\}, 0, 4]\ [0, 3, \{S \to ((c)A(b))s, S \to (A(d))s, S \to (B)s\}, 4]\ [0, 1, \{A \to a\}, 2]$ | 1a |
| $[-1, \{S' \to \bot(S)\}, 0, 4]\ [0, 3, \{S \to ((c)A(b))s, S \to (A(d))s, S \to (B)s\}, 4]\ [1, A \to a, 2]$ | 1d |
| $[-1, \{S' \to \bot(S)\}, 0, 4]\ [0, 3, \{S \to ((c)A(b))s, S \to (A(d))s, S \to (B)s\}, 4]\ [0, 1, \{(c)A(b), A(d), A(b)\}, 2, 3]$ | 7b |
| $[\ \ldots\ ]\ [0, 3, \{S \to ((c)A(b))s, S \to (A(d))s, S \to (B)s\}, 4]\ [0, 1, \{(c)A(b), A(d), A(b)\}, 2, 3]\ [2, 2, \{b\}, 3, 3]$ | 2a |
| $[\ \ldots\ ]\ [0, 3, \{S \to ((c)A(b))s, S \to (A(d))s, S \to (B)s\}, 4]\ [0, 1, \{(c)A(b), A(d), A(b)\}, 2, 3]\ [2, b, 3]$ | 1a, 1d |
| $[\ \ldots\ ]\ [0, 3, \{S \to ((c)A(b))s, S \to (A(d))s, S \to (B)s\}, 4]\ [0, 1, \{(c)A(b), A(b)\}, 3]$ | 4a |
| $[\ \ldots\ ]\ [0, 3, \{S \to ((c)A(b))s, S \to (A(d))s, S \to (B)s\}, 4]\ [0, 1, \{(c)A(b), A(b)\}, 3]\ [0, 0, \{c\}, 1, 1]$ | 3b |
| $[-1, \{S' \to \bot(S)\}, 0, 4]\ [0, 3, \{S \to ((c)A(b))s, S \to (A(d))s, S \to (B)s\}, 4]\ [0, 1, \{(c)A(b), A(b)\}, 3]\ [0, c, 1]$ | 1a, 1d |
| $[-1, \{S' \to \bot(S)\}, 0, 4]\ [0, 3, \{S \to ((c)A(b))s, S \to (A(d))s, S \to (B)s\}, 4]\ [0, (c)A(b), 3]$ | 5b |
| $[-1, \{S' \to \bot(S)\}, 0, 4]\ [0, S \to ((c)A(b))s, 4]$ | 5b |
| $[-1, \{S' \to \bot(S)\}, 0, 4]\ [0, 0, \{S\}, 4, 4]$ | 7a |
| $[-1, \{S' \to \bot(S)\}, 0, 4]\ [0, S, 4]$ | 1a, 1d |
| $[-1, S' \to \bot(S), 4]$ | 5a |

Figure 1: Generalized HI parsing

3a $[k, Q, m, j] \mapsto [k, Q, m, j][m, p-1, Q', p, j]$
  where there is $p$ such that $m < p \leq j$ and $Q' = goto(right(Q), a_p)$ is not empty

3b $[i, k, Q, m] \mapsto [i, k, Q, m][i, p-1, Q', p, k]$
  where there is $p$ such that $i < p \leq k$ and $Q' = goto(left(Q), a_p)$ is not empty

4a $[i, k, Q, m, j][k', \gamma, m'] \mapsto [i, k, Q', m']$
  provided $m = k'$, where $Q' = gotoright(Q, \gamma)$

4b Symmetric to 4a (cf. 2a and 2b)

5a $[k, Q, m, j][k', \gamma, m'] \mapsto [k, t, m']$
  provided $m = k'$, where $t \in gotoright(Q, \gamma)$

5b Symmetric to 5a (cf. 3a and 3b)

6a $[i, k, Q, m, j][k', A \to \gamma, m'] \mapsto$
  $[i, k, Q, m, j][m, k', Q', m', j]$
  provided $m \leq k'$, where $Q' = goto(right(Q), A)$

6b Symmetric to 6a

7a $[k, Q, m, j][k', A \to \gamma, m'] \mapsto$
  $[k, Q, m, j][m, k', Q', m', j]$
  provided $m \leq k'$, where $Q' = goto(right(Q), A)$

7b Symmetric to 7a

The algorithm above is based on the transformation $\tau_{head}$. It is therefore not surprising that this algorithm is reminiscent of LR parsing [1] for a transformed grammar $\tau_{two}(G)$. For most clauses, a rough correspondence with actions of LR parsing can be found: Clauses 2 and 3 correspond with shifts. Clause 5 corresponds with reductions with rules of the form $[X\alpha] \to X\ [\alpha]$ in $\tau_{two}(G)$. Clauses 6 and 7 correspond with reductions with rules of the form $A \to X\ [\alpha]$ in $\tau_{two}(G)$. For Clauses 1 and 4, corresponding actions are hard to find, since these clauses seem to be specific to generalized head-driven parsing.

The reason that we based Algorithm 6 on $\tau_{head}$ is twofold. Firstly, the algorithm above is more appropriate for presentational purposes than an alternative algorithm we have in mind which is not based on $\tau_{head}$, and secondly, the resulting parsers need less sets $Q$. This is similar in the case of LR parsing.[5]

**Example 1** Consider the generalized head grammar with the following rules:

$$S \to ((c)A(b))s \mid (A(d))s \mid (B)s$$
$$A \to a$$
$$B \to A(b)$$

Assume the input is given by $a_1 a_2 a_3 a_4 = c\ a\ b\ s$. The steps performed by the algorithm are given in Figure 1.  □

Apart from HI parsing, also TD, HC, PHI, and EHI parsing can be adapted to generalized head-driven parsing.

## Correctness

The head-driven stack automata studied so far differ from one another in their degree of nondeterminism. In this section we take a different perspective. For all these devices, we show that quite similar relations exist between stack contents and the way input strings are visited. Correctness results easily follow from such characterisations. (Proofs of statements in this section are omitted for reasons of space.)

Let $G = (N, T, P, S)$ be a head grammar. To be used below, we introduce a special kind of derivation.

---

[5]It is interesting to compare LR parsing for a context-free grammar $G$ with LR parsing for the transformed grammar $\tau_{two}(G)$. The transformation has the effect that a reduction with a rule is replaced by a cascade of reductions with smaller rules; apart from this, the transformation does not affect the global run-time behaviour of LR parsing. More serious are the consequences for the size of the parser: the required number of LR states for the transformed grammar is smaller [9].

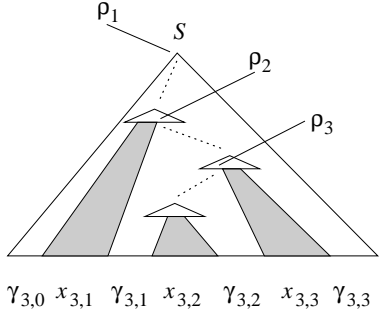

$\gamma_{3,0} \; x_{3,1} \; \gamma_{3,1} \; x_{3,2} \; \gamma_{3,2} \; x_{3,3} \; \gamma_{3,3}$

Figure 2: A head-outward sentential form derived by the composition of $\sigma$-derivations $\rho_i$, $1 \leq i \leq 3$. The starting place of each $\sigma$-derivation is indicated, each triangle representing the application of a single production.

**Definition 1** *A $\sigma$-derivation has the form*

$$\begin{aligned} A &\stackrel{p_1 p_2 \cdots p_{s-1}}{\longrightarrow} & \gamma_0 B \gamma_1 \\ &\stackrel{p_s}{\longrightarrow} & \gamma_0 \alpha \eta \beta \gamma_1 \\ &\stackrel{\rho}{\longrightarrow} & \gamma_0 \alpha x \beta \gamma_1, \end{aligned} \quad (1)$$

*where $p_1, p_2, \ldots, p_s$ are productions in $P^\dagger$, $s \geq 1$, $p_i$ rewrites the unique nonterminal occurrence introduced as the head element of $p_{i-1}$ for $2 \leq i \leq s$, $p_s = (B \to \alpha \underline{\eta} \beta)$ and $\rho \in P^*$ rewrites $\eta$ into $x \in T^+$.*

The indicated occurrence of string $\eta$ in (1) is called the *handle* of the $\sigma$-derivation. When defined, the rightmost (leftmost) nonterminal occurrence in $\alpha$ ($\beta$, respectively) is said to be *adjacent* to the handle. The notions of handle and adjacent nonterminal occurrence extend in an obvious way to derivations of the form $\phi A \theta \stackrel{\rho}{\to} \phi \gamma_0 x \gamma_1 \theta$, where $A \stackrel{\rho}{\to} \gamma_0 x \gamma_1$ is a $\sigma$-derivation.

By composing $\sigma$-derivations, we can now define the class of sentential forms we are interested in. (Figure 2 shows a case example.)

**Definition 2** *A head-outward sentential form is obtained through a derivation*

$$\begin{aligned} S &\stackrel{\rho_1}{\to} & \gamma_{1,0} x_{1,1} \gamma_{1,1} \\ &\stackrel{\rho_2}{\to} & \gamma_{2,0} x_{2,1} \gamma_{2,1} x_{2,2} \gamma_{2,2} \\ &\cdots & \\ &\stackrel{\rho_q}{\to} & \gamma_{q,0} x_{q,1} \gamma_{q,1} x_{q,2} \gamma_{q,2} \cdots \gamma_{q,q-1} x_{q,q} \gamma_{q,q} \end{aligned} \quad (2)$$

*where $q \geq 1$, each $\rho_i$ is a $\sigma$-derivation and, for $2 \leq i \leq q$, only one string $\gamma_{i-1,j}$ is rewritten by applying $\rho_i$ at a nonterminal occurrence adjacent to the handle of $\rho_{i-1}$.*

Sequence $\rho_1, \rho_2, \ldots, \rho_q$ is said to *derive* the sentential form in (2).

The definition of head-outward sentential form suggests a corresponding notion of *head-outward derivation*. Informally, a head-outward derivation proceeds by recursively expanding to a terminal string first the head of a rule, and then the remaining members of the rhs, in an outward order. Conversely, we have *head-inward (HI) derivations*, where first the remaining members in the rhs are expanded, in an inward order (toward the head), after which the head itself is recursively expanded. Note that HI parsing recognizes a string by computing an HI derivation in reverse (cf. LR parsing).

Let $w = a_1 a_2 \cdots a_n$, $n \geq 1$, be a string over $T$ and let $a_0 = \bot$. For $-1 \leq i < j \leq n$, we write $(i,j]_w$ to denote substring $a_{i+1} \cdots a_j$.

**Theorem 1** *For $A$ one of $A^{HC}$, $A^{PHI}$ or $A^{EHI}$, the following facts are equivalent:*

(i) *$A$ reaches a configuration whose stack contents are $I_1 I_2 \cdots I_q$, $q \geq 1$, with*

$$\begin{aligned} I_t &= [i_t, k_t, A_t \to \alpha_t \bullet \eta_t \bullet \beta_t, m_t, j_t] \text{ or} \\ I_t &= [i_t, k_t, A_t \to \eta_t, m_t, j_t] \text{ or} \\ I_t &= [i_t, k_t, \Delta_t \to \eta_t, m_t, j_t] \end{aligned}$$

*for the respective automata, $1 \leq t \leq q$;*

(ii) *a sequence of $\sigma$-derivations $\rho_1, \rho_2, \ldots, \rho_q$, $q \geq 1$, derives a head-outward sentential form*

$$\gamma_0 (k_{\pi(1)}, m_{\pi(1)}]_w \gamma_1 (k_{\pi(2)}, m_{\pi(2)}]_w \gamma_2 \cdots \\ \cdots \gamma_{q-1} (k_{\pi(q)}, m_{\pi(q)}]_w \gamma_q$$

*where $\pi$ is a permutation of $\{1, \ldots, q\}$, $\rho_t$ has handle $\eta_t$ which derives $(k_{\pi(t)}, m_{\pi(t)}]_w$, $1 \leq t \leq q$, and $m_{\pi(t-1)} \leq k_{\pi(t)}$, $2 \leq t \leq q$.*

As an example, an accepting stack configuration $[-1, -1, S' \to \bullet \bot S \bullet, n, n]$ corresponds to a $\sigma$-derivation $(S' \to \bot S)\rho$, $\rho \in P^+$, with handle $\bot S$ which derives the head-outward sentential form $\gamma_0(-1, n]_w \gamma_1 = \bot w$, from which the correctness of the head-corner algorithm follows directly.

If we assume that $G$ does not contain any useless symbols, then Theorem 1 has the following consequence. If the automaton at some point has consulted the symbols $a_{i_1}, a_{i_2}, \ldots, a_{i_m}$ from the input string, $i_1, \ldots, i_m$ increasing indexes, then there is a string in the language generated by $G$ of the form $v_0 a_{i_1} v_1 \cdots v_{m-1} a_{i_m} v_m$. Such a statement may be called *correct subsequence property* (a generalization of correct prefix property [8]). Note that the order in which the input symbols are consulted is only implicit in Theorem 1 (the permutation $\pi$) but is severely restricted by the definition of head-outward sentential form. A more careful characterisation can be obtained, but will take us outside of the scope of this paper.

The correct subsequence property is enforced by the (top-down) predictive feature of the automata, and holds also for $A^{TD}$ and $A^{HI}$. Characterisations similar to Theorem 1 can be provided for these devices. We investigate below the GHI automaton.

For an item $I \in I^{GHI}$ of the form $[i, k, Q, m, j]$, $[k, Q, m, j]$, $[i, k, Q, m]$ or $[k, t, m]$, we say that $k$ ($m$ respectively) is its *left* (*right*) component. Let $N'$ be the set of nonterminals of the head grammar $\tau_{head}(G)$. We need a function yld from reachable items in $I^{GHI}$ into $(N' \cup T)^*$, specified as follows. If we assume that $(\alpha)X(\beta) \in Q \vee A \to (\alpha)X(\beta) \in Q$ and $t = (\alpha)X(\beta) \vee t = A \to (\alpha)X(\beta)$, then

$$\text{yld}(I) = \begin{cases} X & \text{if } I = [i, k, Q, m, j] \\ [\alpha]X & \text{if } I = [k, Q, m, j] \\ X[\beta] & \text{if } I = [i, k, Q, m] \\ [\alpha]X[\beta] & \text{if } I = [k, t, m] \end{cases}$$

It is not difficult to show that the definition of yld is consistent (i.e. the particular choice of a tree or rule from $Q$ is irrelevant).

**Theorem 2** *The following facts are equivalent:*

(i) $A^{GHI}$ *reaches a configuration whose stack contents are* $I_1 I_2 \cdots I_q$, $q \geq 1$, *with* $k_t$ *and* $m_t$ *the left and right components, respectively, of* $I_t$, *and* $\text{yld}(I_t) = \eta_t$, *for* $1 \leq t \leq q$;

(ii) *a sequence of $\sigma$-derivations* $\rho_1, \rho_2, \ldots, \rho_q$, $q \geq 1$, *derives in* $\tau_{head}(G)$ *a head-outward sentential form*

$$\gamma_0 (k_{\pi(1)}, m_{\pi(1)}]_w \gamma_1 (k_{\pi(2)}, m_{\pi(2)}]_w \gamma_2 \cdots$$
$$\cdots \gamma_{q-1} (k_{\pi(q)}, m_{\pi(q)}]_w \gamma_q$$

*where $\pi$ is a permutation of $\{1, \ldots, q\}$, $\rho_t$ has handle $\eta_t$ which derives $(k_{\pi(t)}, m_{\pi(t)}]_w$, $1 \leq t \leq q$, and $m_{\pi(t-1)} \leq k_{\pi(t)}$, $2 \leq t \leq q$.*

## Discussion

We have presented a family of head-driven algorithms: TD, HC, PHI, EHI, and HI parsing. The existence of this family demonstrates that head-driven parsing covers a range of parsing algorithms wider than commonly thought.

The algorithms in this family are increasingly deterministic, which means that the search trees have a decreasing size, and therefore simple realizations, such as backtracking, are increasingly efficient.

However, similar to the left-to-right case, this does not necessarily hold for *tabular* realizations of these algorithms. The reason is that the more refined an algorithm is, the more items represent computation of a single subderivation, and therefore some subderivations may be computed more than once. This is called *redundancy*. Redundancy has been investigated for the left-to-right case in [8], which solves this problem for ELR parsing. Head-driven algorithms have an additional source of redundancy, which has been solved for tabular HC parsing in [14]. The idea from [14] can also be applied to the other head-driven algorithms from this paper.

We have further proposed a generalization of head-driven parsing, and we have shown an example of such an algorithm based on LR parsing. Prospects to even further generalize the ideas from this paper seem promising.